\documentclass[conference]{IEEEtran}
\IEEEoverridecommandlockouts
\usepackage{cite}
\usepackage{amsmath,amssymb,amsfonts}
\usepackage{graphicx}
\usepackage{textcomp}
\usepackage{xcolor}

\def\BibTeX{{\rm B\kern-.05em{\sc i\kern-.025em b}\kern-.08em
    T\kern-.1667em\lower.7ex\hbox{E}\kern-.125emX}}

\usepackage[caption=false,font=footnotesize]{subfig}

\usepackage{bm}
\usepackage{multirow}
\usepackage{float}
\usepackage{algorithm,algpseudocode}
\usepackage{booktabs}
\usepackage{color}
\usepackage{enumitem}

\newtheorem{asu}{Assumption}
\newcounter{subassumption}[asu]

\makeatletter
\renewcommand{\p@subassumption}{\theasu}
\makeatother

\usepackage[capitalise]{cleveref}

\begin{document}

\title{Simultaneously Optimizing Weight and Quantizer of Ternary Neural Network using Truncated Gaussian Approximation}

\author{\IEEEauthorblockN{Zhezhi He\IEEEauthorrefmark{1},
Deliang Fan\IEEEauthorrefmark{1}}

\IEEEauthorblockA{\IEEEauthorrefmark{1}Department of Electrical and Computer Engineering,
University of Central Florida, Orlando, 32816 USA \\
}
\textit{(Elliot.He@knighst.ucf.edu; Dfan@ucf.edu)}
}

\maketitle

\begin{abstract}
In the past years, Deep convolution neural network has achieved great success in many artificial intelligence applications. However, its enormous model size and massive computation cost have become the main obstacle for deployment of such powerful algorithm in the low power and resource-limited mobile systems. As the countermeasure to this problem, deep neural networks with ternarized weights (i.e. -1, 0, +1) have been widely explored to greatly reduce model size and computational cost, with limited accuracy degradation. In this work, we propose a novel ternarized neural network training method which simultaneously optimizes both weights and quantizer during training, differentiating from prior works. Instead of fixed and uniform weight ternarization, we are the first to incorporate the thresholds of weight ternarization into a closed-form representation using truncated Gaussian approximation, enabling simultaneous optimization of weights and quantizer through back-propagation training. With both of the first and last layer ternarized, the experiments on the ImageNet classification task show that our ternarized ResNet-18/34/50 only has $\sim$3.9/2.52/2.16\% accuracy degradation in comparison to the full-precision counterparts.
\end{abstract}

\section{Introduction}

Artificial intelligence is nowadays one of the hottest research topics, which has drawn tremendous efforts from various fields in the past couple years. While computer scientists have succeed to develop Deep Neural Networks (DNN) with transcendent performance in the domains of computer vision, speech recognition, big data processing and etc. \cite{lecun2015deep}. The state-of-the-art DNN evolves into structures with larger model size, higher computational cost and denser layer connections \cite{he2016deep,xie2017aggregated,szegedy2016rethinking,huang2017densely}. Such evolution brings great challenges to the computer hardware in terms of both computation and on-chip storage \cite{he2018optimize}, which leads to great research effort on the topics of model compression in recent years, including channel pruning, weight sparsification, weight quantization, etc. 

Weight ternarization, as a special case of weight quantization technique to efficiently compress DNN model, mainly provides three benefits: 1) it converts the floating-point weights into ternary format (i.e., -1, 0, +1), which can significantly reduce the model size by $16\times$. With proper sparse encoding technique, such model compression rate can be further boosted. 2) Besides the model size reduction, the ternarized weight enables elimination of hardware-expensive floating-point multiplication operations, while replacing with hardware friendly addition/subtraction operations. Thus, it could significantly reduce the inference latency. 3) The ternarized weights with zero values intrinsically prune network connections, thus the computations related to those zero weights can be simply skipped. 

In the previous low-bit network qunatization works, such as TTN \cite{li2016ternary}, TTQ \cite{zhu2016trained} and BNN \cite{courbariaux2016binarized}, they do re-train the models' weights but a fixed weight quantizer is used and not properly updated together with other model parameters, which leads to accuracy degradation and slow convergence of training. In this work, we have proposed a network ternarization method which simultaneously update both weights and quantizer (i.e. thresholds) during training, where our contributions can be summarized as:

\begin{itemize}
    \item We propose a fully trainable deep neural network ternarization method that jointly trains the quantizer threshold, layer-wise scaling factor and model weight to achieve minimum accuracy degradation due to model compression.
    
    \item Instead of fixed and uniform weight ternarization, we are the first to incorporate the thresholds of weight ternarization into a closed-form expression using truncated Gaussian approximation, which can be optimized through back-propagation together with network's other parameters. It differentiates from all precious works.
    
    \item We propose a novel gradient correctness optimization method in straight-through-estimator design. It gives better gradient approximation for the non-differentiable staircase ternarization function, which leads to faster convergence speed and higher inference accuracy.
    
    \item In order to validate the effectiveness of our proposed methods, we apply our model ternarization method on ImageNet dataset for object classification task.

\end{itemize}

The rest of this paper is organized as follows. We first give a brief introduction about the related works regarding the topics of model compression. Then the proposed network ternarization method and the applied tricks are explained in details. In the following section, experiments are performed on both small and large scale dataset with various deep neural network structure, to evaluate the effectiveness of our proposed method. After that, the conclusion is drawn in the end.

\section{Related Works}
Recently, model compression on deep convolutional neural network has emerged as one hot topic in the hardware deployment of artificial intelligence. There are various techniques, including network pruning \cite{luo2017thinet}, knowledge distillation \cite{mishra2017apprentice}, weight sparsification \cite{han2015learning}, weight quantization \cite{han2015deep} and etc. \cite{sandler2018mobilenetv2}, to perform network model compression. As one of the most popular technique, weight quantization techniques are widely explored in many related works which can significantly shrink the model size and reduce the computation complexity \cite{he2018optimize}. The famous deep compression technique \cite{han2015deep} adopts the scheme that optimizing weight quantizer using k-means clustering on the pre-trained model. Even though the deep compression technique can achieve barely no accuracy loss with 8-bit quantized weight, its performance on low-bit quantized case is non-ideal. Thereafter, many works are devoted to quantize the model parameters into binary or ternary formats, not only for its extremely model size reduction ($16\times \sim 32\times$), but also the computations are simplified from floating-point multiplication (i.e. \textit{mul}) operations into addition/subtraction (i.e. \textit{add/sub}). BinaryConnect \cite{courbariaux2015binaryconnect} is the first work of binary CNN which can get close to the state-of-the-art accuracy on CIFAR-10, whose most effective technique is to introduce the gradient clipping. After that, both BWN in \cite{rastegari2016xnor} and DoreFa-Net \cite{zhou2016dorefa} show better or close validation accuracy on ImageNet dataset. In order to reduce the computation complexity, XNOR-Net \cite{rastegari2016xnor} binarizes the input tensor of convolution layer which further converts the \textit{add/sub} operations into bit-wise \textit{xnor} and \textit{bit-count} operations. Besides weight binarization, there are also recent works proposing to ternarize the weights of neural network using trained scaling factors \cite{zhu2016trained}. Leng et. al. employ ADMM method to optimize neural network weights in configurable discrete levels to trade off between accuracy and model size \cite{leng2017extremely}. ABC-Net in \cite{lin2017towards} proposes multiple parallel binary convolution layers to improve the network model capacity and accuracy, while maintaining binary kernel. All above discussed aggressive neural network binarization or ternarization methodologies sacrifice the inference accuracy in comparison with the full precision counterpart to achieve large model compression rate and computation cost reduction.

\section{Methodology}

\subsection{Problem Definition}

As for weight quantization of neural networks, the state-of-the-art work \cite{zhang2018lq} typically divides it into two sub-problems: (1) minimizing quantization noise (i.e., Mean-Square-Error) between floating-point weights and quantized weights and (2) minimizing inference error of neural network w.r.t the defined objective function. In this work, instead of optimizing two separated sub-problems, we mathematically incorporate the thresholds of weight quantizer into neural network forward path, enabling simultaneous optimization of weights and thresholds through back-propagation method. In this work, the network optimization problem can be described as obtaining the optimized ternarized weight $\bm{w}^*$ and ternarization threshold $\bm{\delta}^*$:
\begin{equation}
    \bm{w}^*, \bm{\delta}^* =  \underset{\bm{w},\bm{\delta}}{\text{argmin}}~\mathcal{L}(\bm{y},f(\bm{x},\bm{w},\bm{\delta}))
\end{equation}
where $\mathcal{L}$ is the defined network loss function, $\bm{y}$ is the target corresponding to the network input tensor $\bm{x}$, $f(\cdot)$ computes the network output w.r.t the network parameters.
 
\subsection{Trainable ternarization under Gaussian approximation}
In this subsection, we will first introduce the our weight ternarization methodology. Then, our proposed method to incorporate ternarization thresholds into neural network inference path, which makes it trainable through back-propagation, is discussed particularly. 

\begin{figure}[h]
\centering
	\subfloat[processes to only update thresholds $\bm{\delta}$\label{fig_train_flow1}]{%
		\includegraphics[width=0.90\linewidth]{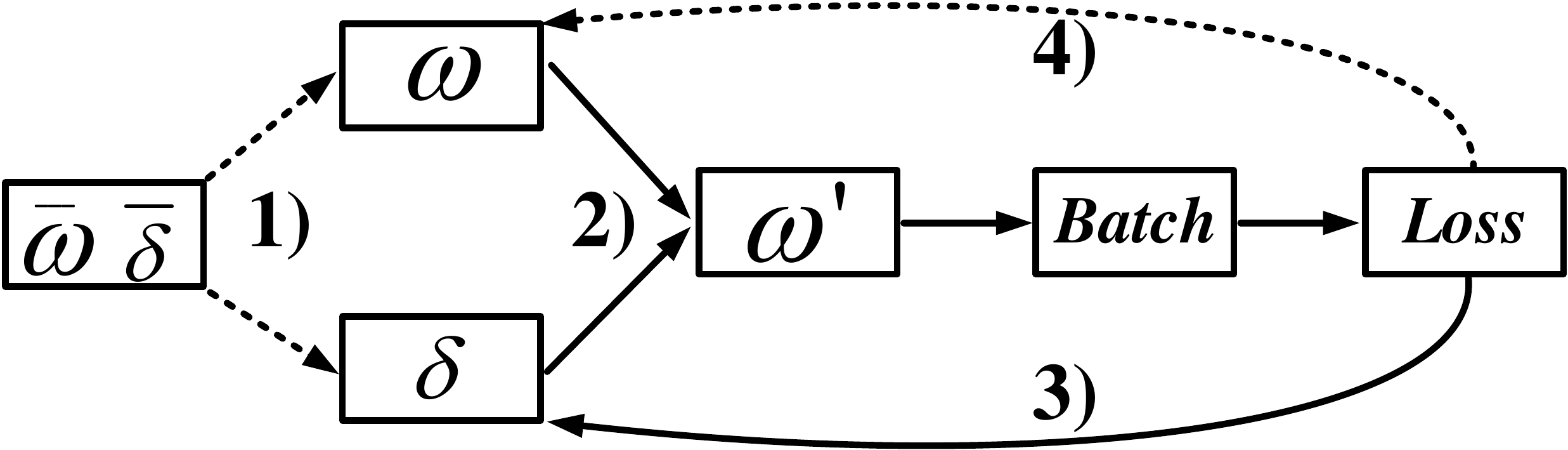}
	} \\ \vspace{-1em}
	\subfloat[processes to only update weights $\bm{w}$\label{fig_train_flow2}]{%
		\includegraphics[width=0.90\linewidth]{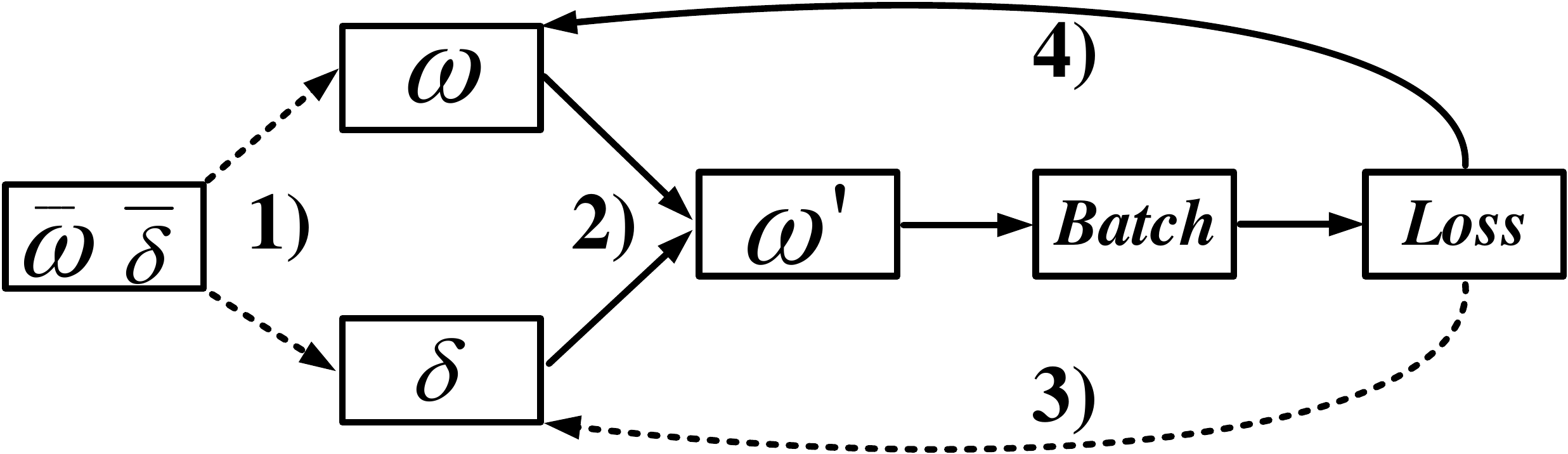}
	}
	\caption{The flowchart of network ternarization, where solid/dashed line indicate activate/inactive step transition. 1)$\Rightarrow$2)$\Rightarrow$3)$\Rightarrow$2)$\Rightarrow$4) steps are sequentially operated during training.}
	\label{fig_train_flow}
\end{figure}

\subsubsection{Network Ternarization:}

For the sake of obtaining a deep neural network with ternarized weight and minimum accuracy loss, the training scheme for one iteration (as shown in \cref{fig_train_flow}) can be generally enumerated as four steps:
\begin{enumerate}[label=\arabic*)]
    \item Initialize the weight with full-precision pre-trained model. Previous works have experimentally demonstrate that fine-tuning the pre-trained model with small learning rate normally generates a quantized model with higher accuracy. More importantly, with the pre-trained model as parameter initialization, much less number of training epochs is required to get model converged in comparison to training from scratch.
    
    \item Ternarize the full-precision weight w.r.t the layer-wise thresholds and scaling factor in real time. The weight ternarization function can be described as:
\begin{equation}
\label{eqt_ternfunc}
\begin{split}
    w_{l,i}' & =S_l(\mu_l,\sigma_l,\Delta_{l}) \cdot Tern(w_{l,i}, \Delta_{l}) \\
    & = S_l(\mu_l,\sigma_l,\Delta_{l}) \cdot
    \begin{cases}
    +1         & w_{l,i} > \Delta_l^+\\
    0         &  \Delta_{l}^- \leq w_{l,i} \leq \Delta_{l}^+\\
    -1         & w_{l,i} < \Delta_{l}^-\\
    \end{cases} 
\end{split}
\end{equation}
where $w_l$ and $w_l'$ denote the full-precision weight base and its ternarized version of $l$-th layer respectively. $\Delta_{l}$ are the ternarization thresholds. $S_l(\cdot)$ calculates the layer-wise scaling factor using extracted mean $\mu_l$, standard deviation $\sigma_l$ and thresholds $\Delta_{l}$, which is the key to enable threshold training in our work. The closed-form expression of $S_l$ will be described here-in-after.
    
    \item For one input batch, this step only adjusts the thresholds through back propagation in a layer-wise manner, meanwhile suspending the update of weight. 
    
    \item For the identical input batch, it repeats step-2 to synchronize the ternarized weight base w.r.t the updated thresholds in step-3. It then disables the update of thresholds and only allows full-precision weight base to be updated. Since the staircase ternarization function ($Tern(\cdot)$ in \cref{eqt_ternfunc}) is non-differentiable owing to its zero derivatives almost everywhere, we adopt the method of Straight-Through-Estimator \cite{bengio2013estimating} similar as previous network quantization works \cite{zhu2016trained}. It is noteworthy that we propose a new gradient correctness algorithm in STE which is critical to improve the convergence speed for weight retraining (see details in following subsections).
    
\end{enumerate}

With ternarized weights, the major computation \footnote{For simplicity, we neglect the bias term. } (i.e., floating-point multiplication and accumulation) is converted to more efficient and less complex floating-point addition and subtraction, due to $Tern(\bm{w}_{l,i})\in\{-1,0,+1\}$ . The computation can be expressed as:
\begin{equation}
\label{eqt_reformat_dotproduct}
     \bm{x}_{l}^T\cdot \bm{w}_{l}' = \bm{x}_{l}^T\cdot (S_l \cdot Tern(\bm{w}_l)) = S_l \cdot( \bm{x}_{l}^T \cdot Tern(\bm{w}_l) )
\end{equation}
where $\bm{x}_{l}$ and $\bm{w}_{l}'$ are the vectorized input and ternarized weight of layer $l$ respectively. 
Since in the structures of state-of-the-art deep neural networks, convolution/fully-connected layers normally follows a batch-normalization layer or ReLU where both layers perform element-wise function on their input tensor (i.e., $\bm{x}_{l}^T\cdot \bm{w}_{l}'$), the element-wise scaling in \cref{eqt_reformat_dotproduct} can be emitted and integrate with the forthcoming batch-norm and ReLU. Beyond the above descriptions of ternarized model training procedure, we formalize those operations in \cref{alg:algorithm1} as well for clarification.

\begin{figure*}[h]
	\centering
		\begin{tabular}{l}
		\includegraphics
		[width=0.98\textwidth]{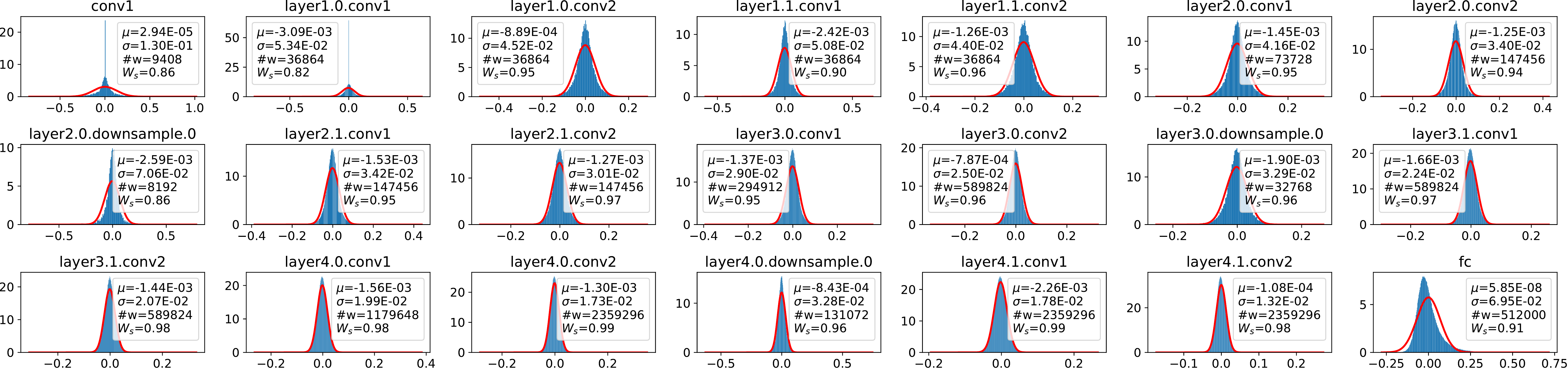}\\
		\end{tabular}
    	\caption{the histogram of weights (blue shadow) along with the Gaussian distribution PDF curve (red line) w.r.t extracted mean $\mu$ and $\sigma$ for convolution, fully-connected and residual layers in ResNet-18b \cite{he2016deep}. For layers with more number of weights ($\# w$), the weights distribution conforms to the Gaussian distribution more precisely. }
	\label{fig_distribution_exam}
\end{figure*}

\subsubsection{Trainable thresholds utilizing truncated Gaussian distribution approximation:}

It has been fully discussed in previous works \cite{han2015deep, blundell2015weight} that the weight distributions of spatial convolution layers and fully-connected layers are intending to follow Gaussian distribution, whose histogram is in bell-shape, owing to the effect of $\mathcal{L}2$-regularization. For example, in \cref{fig_distribution_exam}, we have shown the weight distributions and their corresponding Probability Density Function (PDF) using the calculated mean and standard deviation for each parametric layer (i.e., convolution and fully-connected layers) in ResNet-18b \cite{he2016deep}. Meanwhile, the Shapiro-Wilk normality test \cite{shapiro1965analysis} is conducted to identify whether the weight sample originated from Gaussian distribution quantitatively. The given test statistic $ \mathcal{W}_s$ of Shapiro-Wilk normality test indicate a good normally distribution match with minimum 0.82 value. Note that, the asymmetry (i.e., Skewness) of the last fully-connected layer is due to the existence of bias term. In this work, we consider the weight of parametric layers approximately following Gaussian distribution and further perform the weight ternarization based on such approximation.


In order to make the thresholds as trainable parameters that can be updated using back-propagation, there are two criteria that have to be met:
\begin{itemize}
    \item thresholds have to be parameters of inference path in a closed-form expression.
    \item such closed-form expression is differentiable w.r.t the threshold.
\end{itemize}
Hereby, we make the assumption that:
\begin{asu}
the weights of designated layer $l$ are approximately following Gaussian distribution (i.e., $\bm{w}_l\sim \mathcal{N}(\mu_l, \sigma^2_l)$), where $\mu_l$ and $\sigma_l$ are the calculated mean and standard deviation of the weight sample $\bm{w_l}$.
\end{asu}
The centroid is normally taken as the quantized value for a nonuniform quantizer setup to minimize the quantization error \cite{proakis1994communication}. Thus, for weight ternarization, the layer-wise scaling factor can be set as:
\small
\begin{equation}
\label{eqt_scaling_old}
\begin{split}
    S_l(\bm{w_l}, \Delta_l) & =  \int_{-\infty}^{\Delta_{l}^-}\phi_c(x) \cdot xdx
    +\int_{\Delta_{l}}^{+\infty}\phi_c(x)\cdot x dx \\
    & = E(|\bm{w}_{l,i}| \big \vert (\bm{w}_{l,i}>\Delta_{l}^+) \cup (\bm{w}_{l,i}<\Delta_{l}^-) )
\end{split}
\end{equation}
where $\phi_c(x)$ is the conditional PDF under the condition of $(x>\Delta_l^+) \lor (x<\Delta_l^-)$. In this work, by setting $\Delta_l^+ = \mu_l + \delta_l$ and $\Delta_l^- = \mu_l - \delta_l$, we can approximate the \cref{eqt_scaling_old} and reformat it into:
\small
\begin{equation}
\label{eqt_scaling_new}
    S_l(\mu_l,\sigma_l,\delta_{l}) =  \int_{a=\mu_l + \delta_l}^{b=+\infty} \dfrac{\phi(x \big\vert \mu_l,\sigma_l)}{\Phi(b \big\vert \mu_l,\sigma_l)-\Phi(a \big\vert \mu_l,\sigma_l)} \cdot x dx 
\end{equation}
where $\phi(x\big\vert \mu_l,\sigma_l))$ and $\Phi(x\big\vert \mu_l,\sigma_l))$ are the PDF and CDF for Gaussian distribution $\mathcal{N}(\mu_l, \sigma^2_l)$ . Such calculation can directly utilize the closed-form expression of mathematical expectation for truncated Gaussian distribution with lower bound $a$ and upper bound $b$. Thus, we finally obtain a closed-form expression of scaling factor embedding trainable thresholds $\delta_l$:
\begin{equation}
    \alpha = \dfrac{a-\mu_l}{\sigma_l}=\dfrac{\delta_l}{\sigma_l};~~~ \beta= \dfrac{b-\mu_l}{\sigma_l}=+\infty
\end{equation}
\begin{equation}
\begin{split}
    S_l(\mu_l,\sigma_l,\delta_{l}) &= \mu_l - \sigma_l \cdot \dfrac{\phi(\beta|0,1)-\phi(\alpha|0,1)}{\Phi(\beta|0,1)-\Phi(\alpha|0,1)}  \\
    &= \mu_l+\sigma_l\cdot \dfrac{\phi(\alpha|0,1)}{1-\Phi(\alpha|0,1)} 
\end{split}
\end{equation}
where $\phi(\cdot \big \vert 0,1)$ and $\Phi(\cdot \big \vert 0,1)$ are PDF and CDF of standard normal distribution $\mathcal{N}(0, 1)$.

\begin{figure}[h]
\centering
	\subfloat[\label{fig_Sl_curve1}]{%
		\includegraphics[width=0.96\linewidth]{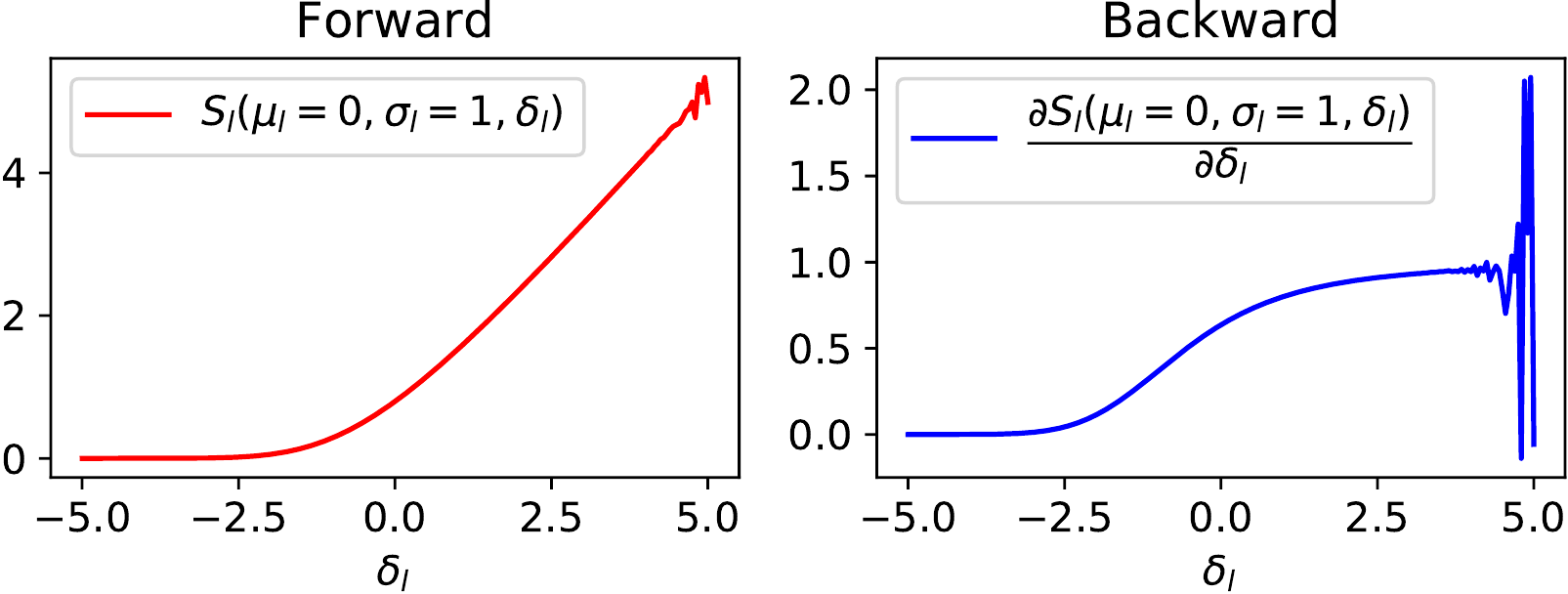}
	} \\ \vspace{-1em}
	\subfloat[\label{fig_Sl_curve2}]{%
		\includegraphics[width=0.96\linewidth]{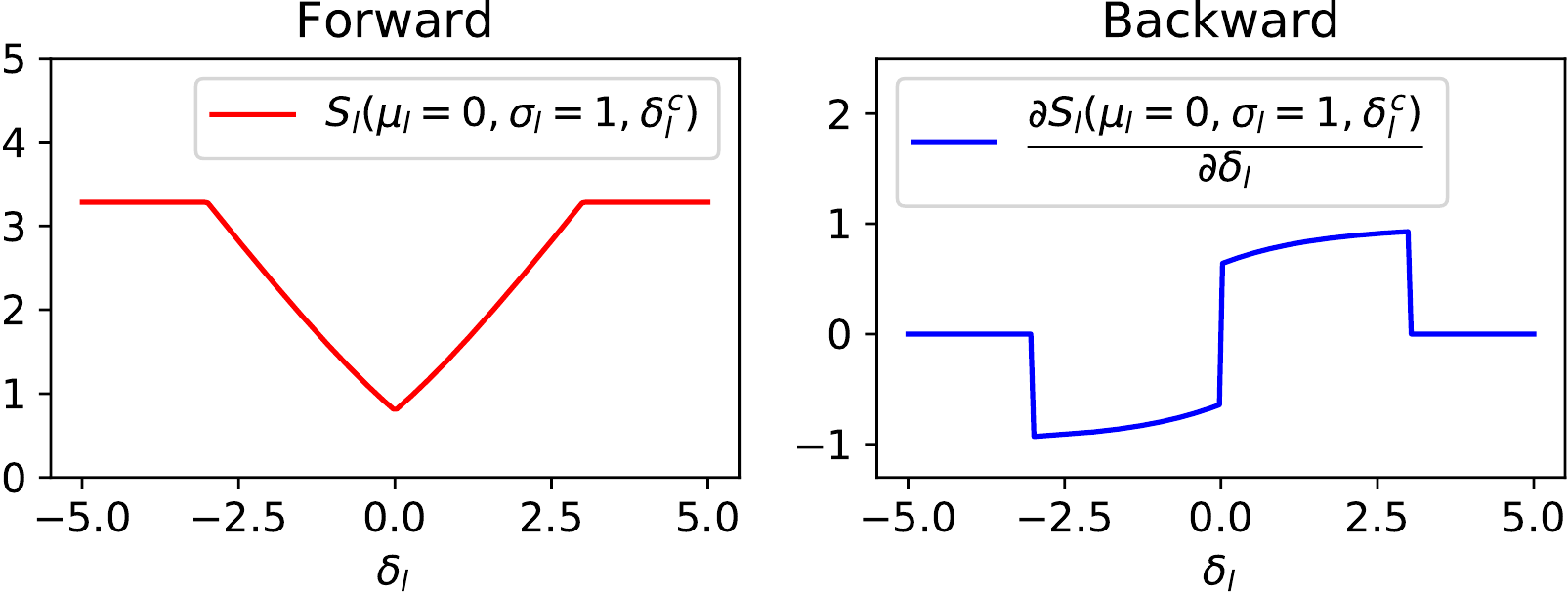}
	}
	\caption{The forward and backward curves for (a) $S_l(\mu_l, \sigma_l, \delta_l)$ and (b) $S_l(\mu_l, \sigma_l, \delta_l^c)$ w.r.t $\delta_l$, where $\delta_l^c$ is $\delta_l$ with clipping constraints . Note that, we choose $\mu_l=0$ and $\sigma_l=1$ as the example for visualization.} 
	\label{fig_Sl_curve}
\end{figure}


For visualization purpose, as shown in \cref{fig_Sl_curve1}, we plot the function of $S_l$ in the forward and backward paths w.r.t the variation of $\delta_l$. Moreover, in order to ensure $(\mu_l-\sigma_l) < (\mu_l-\sigma_l)$ for correct ternarization and prevent framework convergence issue \footnote{Since most of the popular deep learning frameworks using numerical method (e.g., Monte-Carlo method) for distribution related calculation, there will be error for calculating $S_l$ and $\partial S_l/\partial \delta_l$ at the tail of distribution (i.e., $\delta_l > 3\times \sigma_l$)}, we apply constraints on $\delta_l$ which keep $ abs(\delta_l)\in(0,3 \sigma_{l})$. Such clipping operation is functionally equivalent as propagating $\delta_l$ through the \textit{hard tanh} function, which is piece-wise linear activation function with upper-limit $j$ and lower-limit $k$, then the trainable thresholds with clipping constraints can be expressed as:
\begin{equation}
    \textup{hardtanh}(x,j,k) = \textup{Clip}(x,j,k) = \textup{max}(j,\textup{min}(x,k))
\end{equation}
\begin{equation}
    \delta_l^c = \textup{hardtanh}(\textup{abs}(\delta_l), 0, 3 \sigma_{l})
\end{equation}
After substituting the vanilla $\delta_l$ with the $\delta_l^c$ \cref{fig_Sl_curve2} in calculating $S_l$, the function of forward ($S_l(\mu_l,\sigma_l,\delta_{l}^c)$) and backward path ($\partial  S_l(\mu_l,\sigma_l,\delta_{l}^c)/\partial \delta_l$) w.r.t $\delta_l$ is transformed into \cref{fig_Sl_curve2}.

Beyond that, the weight decay tends to push the trainable threshold $\delta_l$ close to zero, and biases the ternary weight representation towards the binary counterpart, thus lowering the sparsity. Therefore, we do not apply weight decay on threshold $\delta_l$ during training.

In summary, we finalize the scaling factor term and weight ternarization function to substitute the original full-precision weight in the forward propagation path:
\begin{equation}
\label{eqt_scaling_final}
    S_l(\mu_l,\sigma_l,\delta_{l}) = \mu_l+\sigma_l\cdot \dfrac{\phi(\delta_l^c/\sigma_l|0,1)}{1-\Phi(\delta_l^c/\sigma_l|0,1)}   
\end{equation}
\begin{equation}
\label{eqt_ternfunc_final}
    Tern(w_{l,i},\mu_l,\delta_l) =
        \begin{cases}
    +1         & w_{l,i} > \mu_l+\delta_l^c\\
    0         &  \mu_l-\delta_l^c \leq w_{l,i} \leq \mu_l+\delta_l^c\\
    -1         & w_{l,i} < \mu_l-\delta_l^c\\
    \end{cases}     
\end{equation}


\subsection{Straight Through Estimator with Gradient Correctness}
\begin{figure}[h]
\centering
	\subfloat[\label{fig_sign_ste}]{%
		\includegraphics[width=0.47\linewidth]{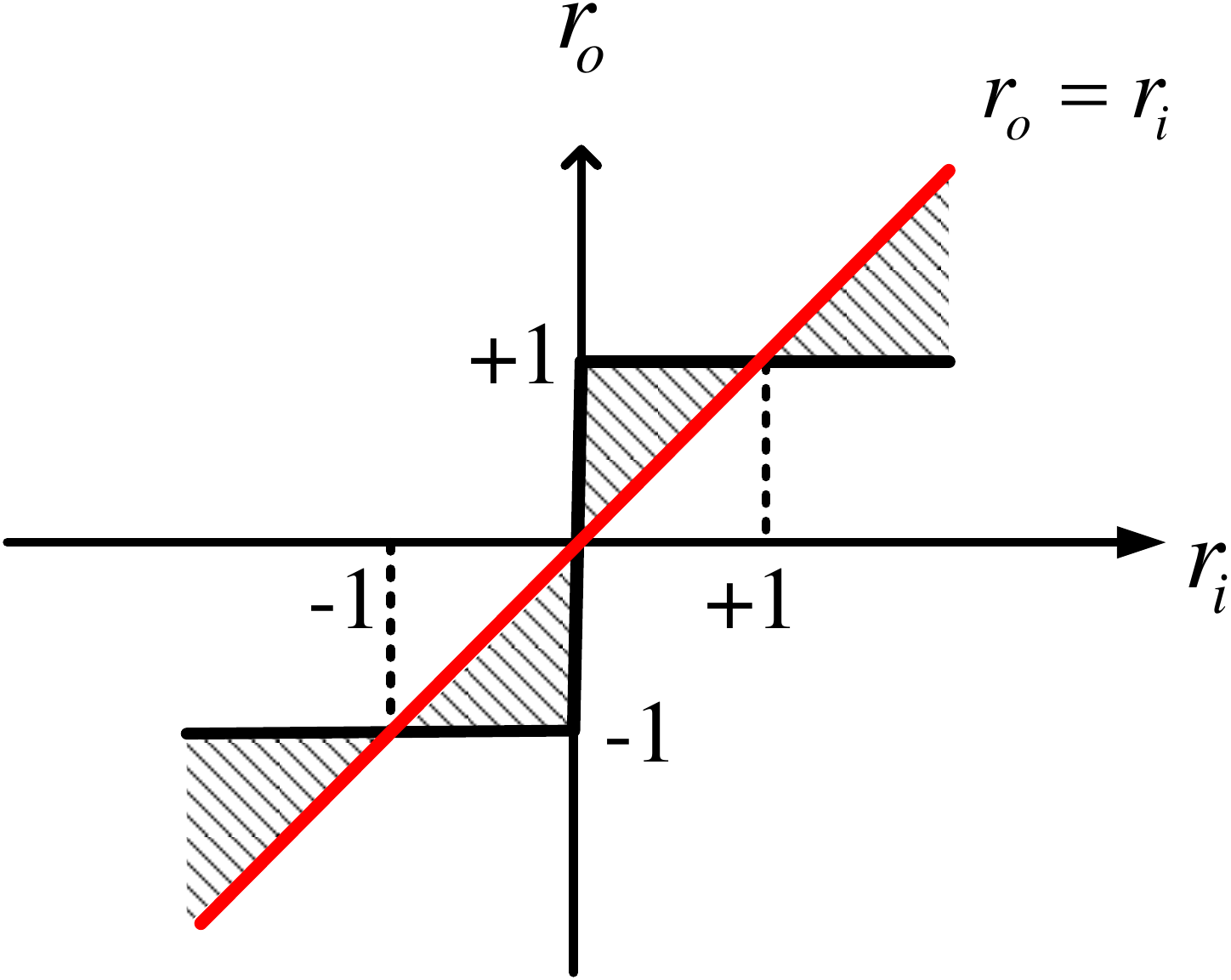}
	}
	\subfloat[\label{fig_tern_ste}]{%
		\includegraphics[width=0.49\linewidth]{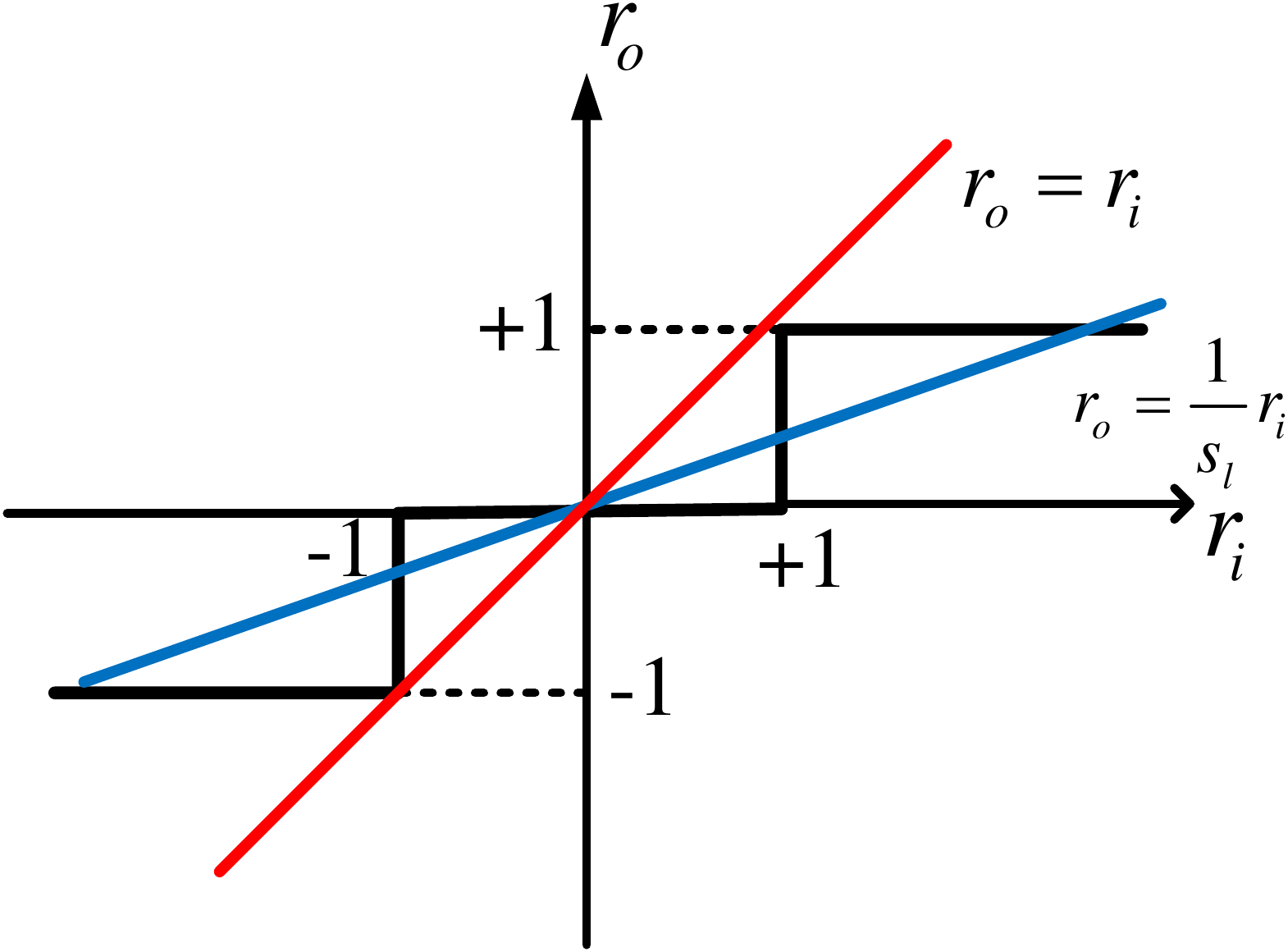}
	}
	\caption{Analysis about the quantizer's straight-through-estimator design for (a) $r_o=sign(r_i)$ for \cite{courbariaux2016binarized} and (b) $r_o=Tern(r_i)$ in this work. } 
	\label{fig_ste_curve}
\end{figure}

Almost for any quantization function which maps the continuous values into discrete space, it has encountered the same problem that such stair-case function is non-differentiable. Thus, a widely adopted countermeasure is using the so-called Straight-Through-Estimator (STE) to manually assign an approximated gradient to the quantization function. We take the STE in famous binarized neural network \cite{courbariaux2016binarized} as an example to perform the analysis, which is defined as:
\begin{equation}
\label{eqt_bnn_ste_forward}
    \textup{\textbf{Forward}}:~~r_o = \textup{Sign}(r_i)
\end{equation}
\begin{equation}
\label{eqt_bnn_ste_backward}
    \textup{\textbf{Backward}}:~~\dfrac{\partial \mathcal{L}}{\partial r_o} \stackrel{\textup{STE}}{=} \dfrac{\partial \mathcal{L}}{\partial r_i}\bigg\vert_{|r_i|\leq 1} \Longrightarrow \dfrac{\partial r_o}{\partial r_i}\bigg\vert_{|r_i|\leq 1}=1
\end{equation}
where $\mathcal{L}$ is the defined loss function.
The rule behind such STE setup is that the output of quantization function $r_o$ can effectively represent the full-precision input value $r_i$. Thus, $\textup{Sign}(\cdot)$ performs the similar function as $f(r_i)=r_i$ whose derivative is $\partial f(r_i)/\partial r_i =1$. However, the rough approximation in \cref{eqt_bnn_ste_forward} and \cref{eqt_bnn_ste_backward} leads to significant quantization error and hamper the network training when $r_i<<1$ or $r_i>>1$. For example, as shown in \cref{fig_sign_ste}, if the layer's weight distribution owns $\sigma_l<<1$, performing network quantization through fine-tuning will result in significant weight distribution shift which slows down the convergence speed.
    
In order to encounter the drawback of naive STE design discussed above, we propose a method called \textit{gradient correctness} for better gradient approximation. For our weight ternarization case, the full-precision weight base $\bm{w}_l$ is represented by $S_l(\mu_l,\sigma_l,\delta_l)\cdot Tern(\bm{w}_l)$, where both terms can pass back gradients to update the embedding parameters. For assigning a proper gradient to the $Tern(w_{l,i})$, we follow STE design rule which leads to the following expression:
\begin{equation}
\label{eqt_STE_equality}
   \dfrac{\partial w_{l,i}'}{\partial w_{l,i}} =  \dfrac{\partial S_l\cdot Tern(w_{l,i})}{\partial w_{l,i}} = S_l \dfrac{\partial Tern(w_{l,i})}{\partial w_{l,i}} = 1
\end{equation}
Thus, the STE for ternarization function can be derived \cref{eqt_STE_equality} as:
\begin{equation}
\label{eqt_STE_scale}
    \dfrac{\partial Tern(w_{l,i})}{\partial w_{l,i}} = \dfrac{1}{S_l}
\end{equation}
As seen in \cref{eqt_STE_scale}, instead of simply assigning the gradient as 1, we scale the $\partial Tern(w_{l,i})/\partial w_{l,i}$ w.r.t the value of $S_l(\mu_l,\sigma_l,\delta_l)$ in real time. As shown in \cref{fig_tern_ste}, STE could better approximate the gradient with adjustable gradient correctness term.

\begin{algorithm}[ht]
  \caption{Training both the weights and thresholds of ternarized network under the assumption that weights are following Gaussian distribution. }
  \label{alg:algorithm1}
  \begin{algorithmic}[1]
    \Require: a mini-batch of inputs $\bm{x}$ and its corresponding targets $\bm{y}_t$, number of layers $N$, full-precision pretrained weights $\bar{\bm{w}}$, initial thresholds $\bm{\delta}$ full-precision weight base $\bm{w}^t$ and layer-wise thresholds $\bm{\delta}^{t}$ from last training iteration $t$, learning rate $\eta$, network inference function $f(\cdot)$.
    \Ensure for current iteration index of $t+1$, updated full-precision weights $\bm{w}^{t+1}$, updated layer-wise thresholds $\bm{\delta}^{t+1}$.
    \Statex \{Step-1. Initialization:\}
    \If{$t = 0$} \Comment{This is the first training iteration}     \State $\bm{w} \gets \bar{\bm{w}}; \bm{\delta} \gets \bar{\bm{\delta}}$ \Comment{load pretrained model} 
    \Else 
         \State $\bm{w} \gets \bm{w}^{t}; \bm{\delta} \gets \bm{\delta}^t$ \Comment{load from last iteration}
    \EndIf
    \Statex \{Step-2. Weight ternarization:\}
    \For{$l$ := $1$ to $N$}
        \State $\mu_l \gets \bm{w}_l.\textup{mean()}; \sigma_l \gets \bm{w}_l.\textup{std()}$
        \State $\bm{w}_l' \gets S_l(\mu_l, \sigma_l, \delta_l)\cdot Tern(\bm{w}_l,\mu_l,\delta_l)$ \Comment{\cref{eqt_scaling_final,eqt_ternfunc_final}}
    \EndFor
    
    \Statex \{Step-3. Update thresholds $\bm{\delta}$ only:\}
    \State $\bm{y} \gets f(\bm{x},\bm{w'})$
    \Comment{forward propagation, \cref{eqt_reformat_dotproduct}}
    \State $\mathcal{L} \gets Loss(\bm{y},\bm{y}_t)$
    \Comment{get inference error}
    \For{$l$ := $N$ to $1$}
        \State $g_{\delta_l} \gets \partial\mathcal{L}/\partial \delta_l $
        \Comment{back-propagate for gradients}
        \State $\delta_l \gets   \textup{Update}(\delta_l, g_{\delta_l},\eta )$
        \Comment{Using vanilla SGD}
    \EndFor
    
    \Statex \{Repeat Step-2: from op-6 to op-10\} 
    \Comment{important step!}
        
    \Statex \{Step-4. Update weights $\bm{w}$ only:\}
    \State $\bm{y} \gets f(\bm{x},\bm{w'})$
    \State $\mathcal{L} \gets Loss(\bm{y},\bm{y}_t)$
    \For{$l$ := $N$ to $1$}
        \State $g_{\bm{w}_l} \gets \partial\mathcal{L}/\partial \bm{w}_l $
        \Comment{back-propagate for gradients}
        \State $\bm{w}_l \gets   \textup{Update}(\bm{w}_l, g_{\bm{w}_l},\eta )$
        \Comment{Using SGD/Adam}
    \EndFor
    
    \noindent\Return $\bm{w}^{t+1} \gets \bm{w}; \bm{\delta}^{t+1} \gets \bm{\delta}  $
    
  \end{algorithmic}
\end{algorithm}

\section{Experiment and Result Evaluation}

\subsection{Experiment setup}
In this work, we evaluate our proposed network ternarization method for object classification task with CIFAR-10 and ImageNet datasets. All the experiments are performed under Pytorch deep learning framework using 4-way Nvidia Titan-XP GPUs. For clarification, in this work, both the first and last layer are ternarized during the training and test stage.

\subsubsection{CIFAR-10:}
CIFAR-10 contains 50 thousands training samples and 10 thousands test samples with $32\times 32$ image size. The data augmentation method is identical as used in \cite{he2016deep}. For fine-tuning, we set the initial learning rate as 0.1, which is scheduled to scale by 0.1 at epoch 80, 120 respectively. The mini-batch size is set to 128.

\subsubsection{ImageNet:}
In order to provide a more comprehensive experimental results on large dataset, we examine our model ternarization techniques on image classification task with ImageNet \cite{russakovsky2015imagenet} (ILSVRC2012) dataset. ImageNet contains 1.2 million training images and 50 thousands validation images, which are labeled with 1000 categories. For the data pre-processing, we choose the scheme adopted by ResNet \cite{he2016deep}. Augmentations applied to the training images can be sequentially enumerated as: $224\times 224$ randomly resized crop, random horizontal flip, pixel-wise normalization. All the reported classification accuracy on validation dataset is single-crop result. The mini-batch size is set to 256.

\subsection{Ablation studies}
In order to exam the effectiveness of our proposed methods, we have performed the following ablation studies. The experiments are conducted with ResNet-20 \cite{he2016deep} on CIFAR-10 dataset, where the differences are significant enough to tell the effectiveness.

\begin{table}[ht]
\centering
\caption{Ablation study of proposed method using ResNet-20 on CIFAR-10 dataset.}
\label{table:ablation}
\begin{tabular}{@{}cc@{}}
\toprule
Configurations & Accuracy \\ \midrule
full-precision (baseline) & 91.7\% \\ 
\midrule
w/ gradient correctness & 90.39\% \\
w/o gradient correctness & 87.89\% \\
\midrule
vanilla SGD & 90.39\%\\
Adam & 56.31\%\\
\midrule
Initialize with  $\delta_l=0.05\textup{max}(|\bm{w}_l|)$
 & 89.96\% \\ 
Initialize with  $\delta_l=0.1\textup{max}(|\bm{w}_l|)$
 & 90.24\% \\ 
Initialize with  $\delta_l=0.15\textup{max}(|\bm{w}_l|)$
 & 90.12\% \\ 
\bottomrule
\end{tabular}
\end{table}

\begin{figure}[t]
	\centering
		\begin{tabular}{l}
		\includegraphics
		[width=0.9\linewidth]{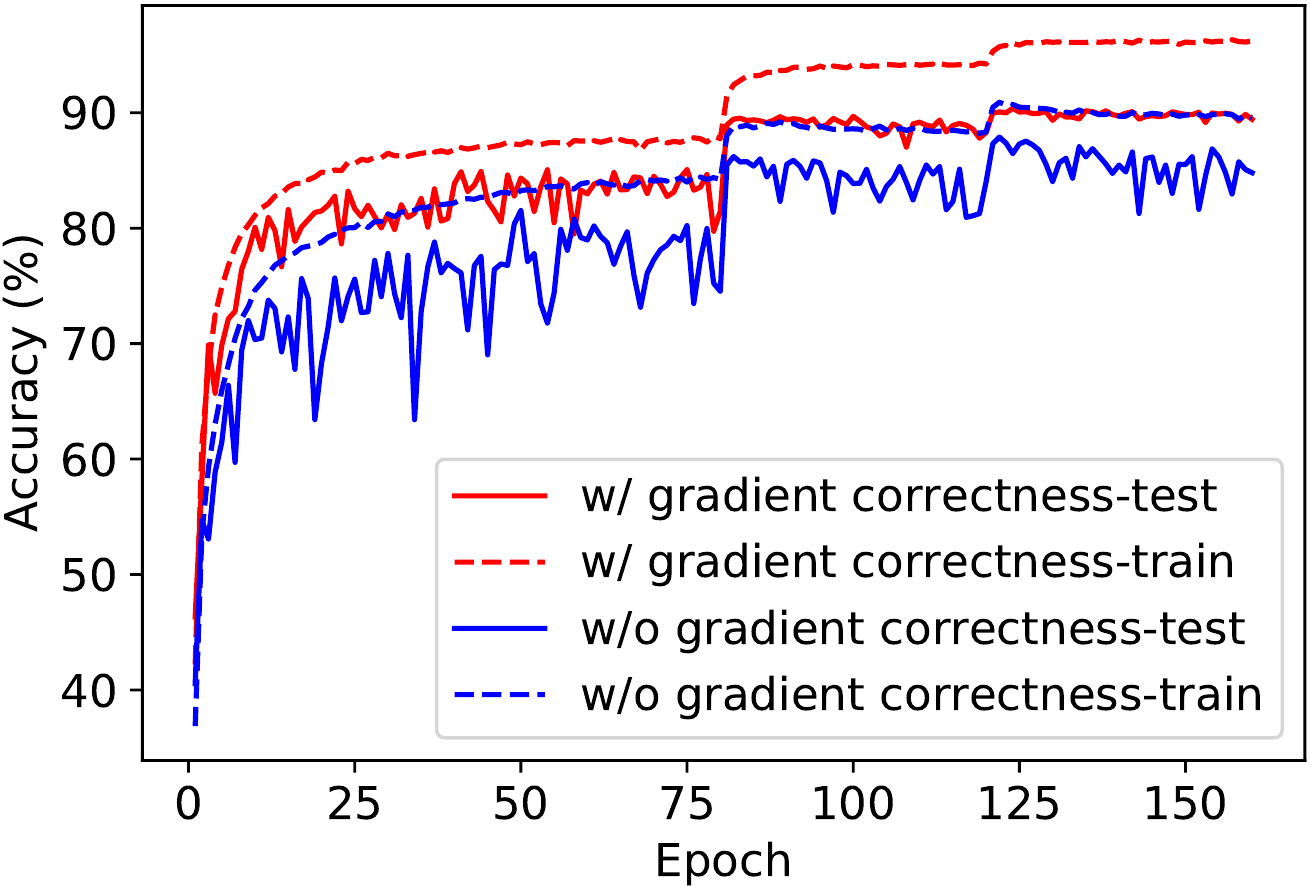}\\
		\end{tabular}
    	\caption{The accuracy evolution curve for train and test for the cases w/ or w/o gradient correctness.}
	\label{fig_ablation_1}
\end{figure}

\subsubsection{Gradient Correctness}
We compare the accuracy curve convergence speed between the STE design with or without the gradient correctness. As shown in \cref{fig_ablation_1}, the network training speed with gradient correctness is much faster in comparison with the case without gradient correctness. The main reason cause the convergence speed degradation is that when layer-wise scaling factor is less than 1, without gradient correctness, the gradient of the loss function w.r.t the weights is scaled by the scaling factor due to the chain-rule. Thus, weights are updated with a much smaller step-size in comparison to the thresholds, when optimizers are set up with identical parameters (e.g., learning rate, etc.).



\begin{table*}[ht]
\centering
\caption{Validation accuracy (top1/top5 \%) of ResNet-18/34/50b \cite{he2016deep} on ImageNet using various model quantization methods.}
\label{table:resnet_imagenet}
\scalebox{1}{
\begin{tabular}{@{}cccccc@{}}
\toprule
     & 
\begin{tabular}[c]{@{}c@{}}Quan.\\scheme\end{tabular}&
\begin{tabular}[c]{@{}c@{}}First\\layer \end{tabular} & 
\begin{tabular}[c]{@{}c@{}}Last\\layer\end{tabular} & 
\begin{tabular}[c]{@{}c@{}}Accuracy\\ (top1/top5)\end{tabular}  & \begin{tabular}[c]{@{}c@{}}Comp.\\ rate\end{tabular} \\
\midrule
\midrule
\multicolumn{6}{c}{\textbf{ResNet-18b}} \\ 
\midrule
Full precision & - & FP & FP & 69.75/89.07 & 1$\times$ \\
BWN\cite{rastegari2016xnor} & Bin. & FP & FP & 60.8/83.0  & $\sim$32$\times$ \\
ABC-Net\cite{lin2017towards} & Bin. & FP* & FP* & 68.3/87.9  &  $\sim$6.4$\times$\\
ADMM\cite{leng2017extremely}& Bin. & FP* & FP* & 64.8/86.2  & $\sim$32$\times$ \\
TWN\cite{li2016ternary,leng2017extremely} & Tern. &  FP & FP  & 61.8/84.2  & $\sim$16$\times$ \\
TTN\cite{zhu2016trained}& Tern. &  FP & FP  & 66.6/87.2  & $\sim$16$\times$ \\
ADMM\cite{leng2017extremely}& Tern. & FP* & FP* & 67.0/87.5  & $\sim$16$\times$ \\
APPRENTICE\cite{mishra2017apprentice}& Tern. & FP* & FP* & 68.5/-  & $\sim$16$\times$ \\
this work & Tern. & Tern & Tern & 66.01/86.78 & $\sim$16$\times$ \\ 
\midrule
\multicolumn{6}{c}{\textbf{ResNet-34b}} \\ 
\midrule
Full precision & - & FP & FP & 73.31/91.42 & 1$\times$ \\
APPRENTICE\cite{mishra2017apprentice}& Tern. & FP* & FP* & 72.8/-  & $\sim$16$\times$ \\
this work & Tern. & Tern & Tern & 70.79/89.89 & $\sim$16$\times$ \\
\midrule
\multicolumn{6}{c}{\textbf{ResNet-50b}} \\ 
\midrule
Full precision & - & FP & FP & 76.13/92.86 & 1$\times$ \\
APPRENTICE\cite{mishra2017apprentice}& Tern. & FP* & FP* & 74.7/-  & $\sim$16$\times$ \\
this work & Tern. & Tern & Tern & 73.97/91.65 & $\sim$16$\times$ \\
\bottomrule
\end{tabular}}
\end{table*}

\subsubsection{Optimizer on thresholds}
\begin{figure}[t]
	\centering
		\begin{tabular}{l}
		\includegraphics
		[width=0.9\linewidth]{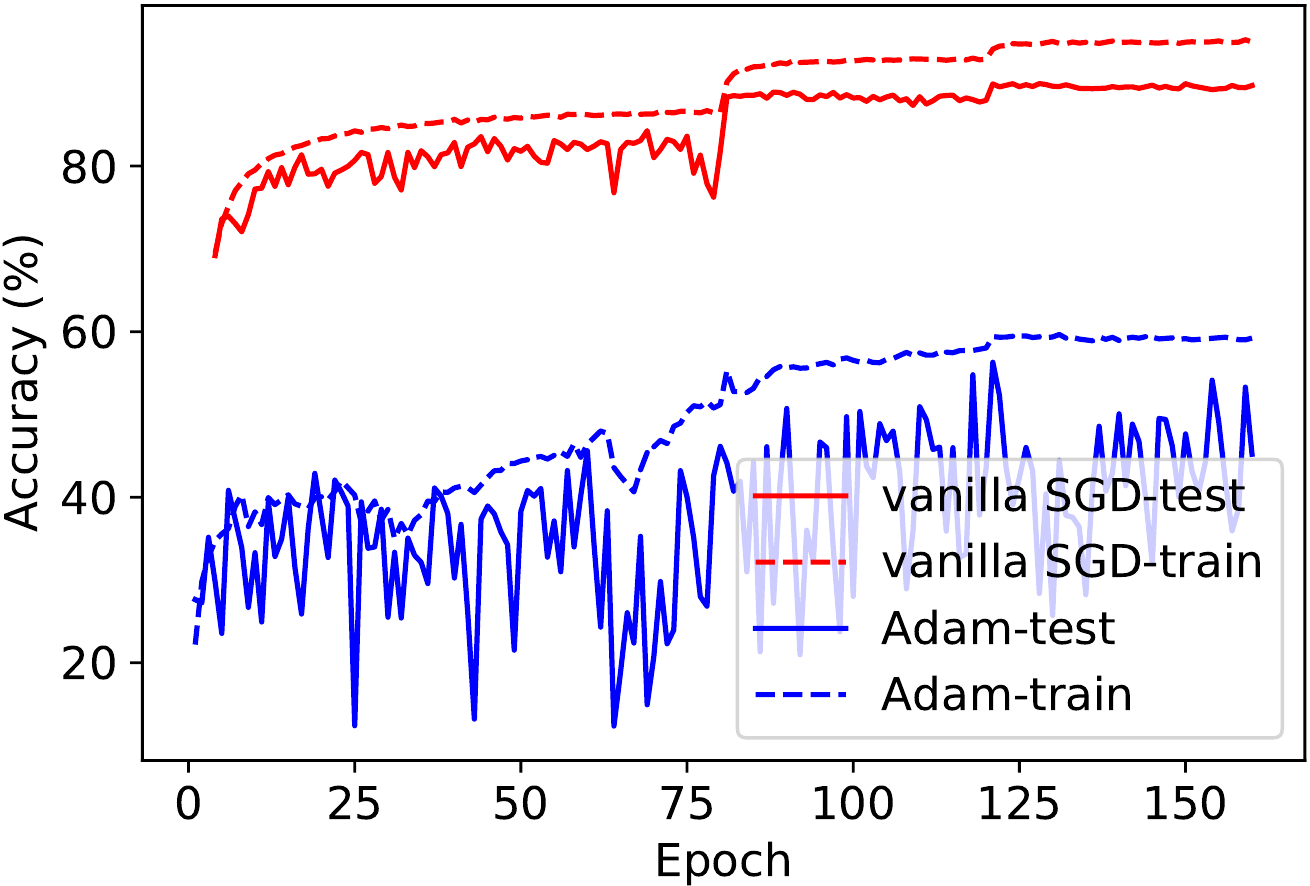}\\
		\end{tabular}
    	\caption{The accuracy evolution curve for train and test for the cases with vanilla SGD and Adam optimizer}
	\label{fig_ablation_3}
\end{figure}

The vanilla SGD and Adam are two most adopted optimizers for quantized neural network training. Hereby, we took those two optimizers as an example to show the training evolution. Note that, since weights and thresholds are iteratively updated for each input mini-batch, we can use different optimizer for weights and thresholds. In this experiment, we use SGD for weight optimization, while using SGD and Adam on thresholds. The result depicted in \cref{fig_ablation_3} shows that it is better to use the same SGD optimizers to achieve higher accuracy. 

\subsubsection{Thresholds Initialization}
\begin{figure}[t]
	\centering
		\begin{tabular}{l}
		\includegraphics
		[width=0.9\linewidth]{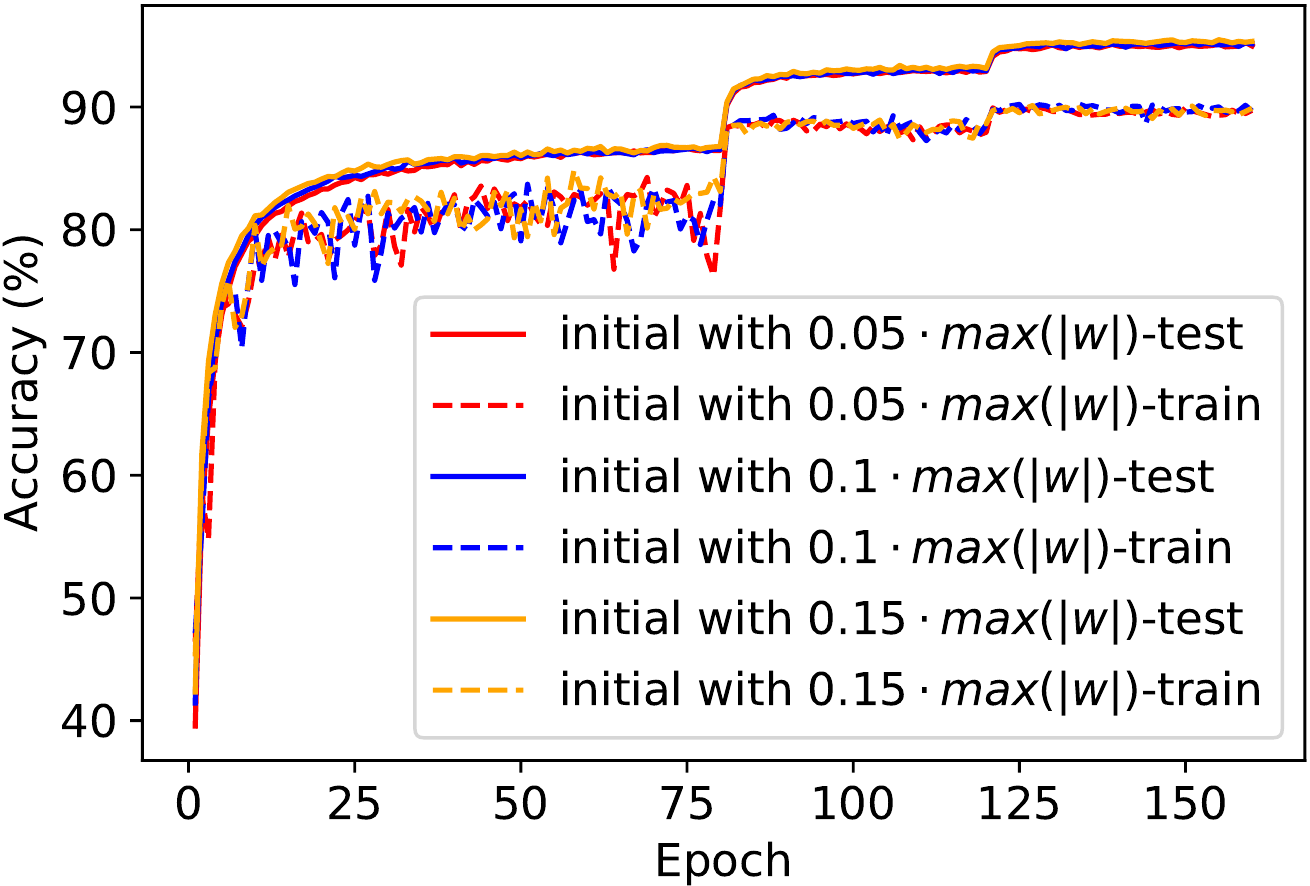}\\
		\end{tabular}
    	\caption{The accuracy evolution curve for train and test for the cases with various threshold initialization.}
	\label{fig_ablation_4}
\end{figure}

In order to exam how the threshold initialization affects the network training, we initialize the threshold as $\delta_l = \{0.05, 0.1, 0.15\}\cdot \textup{max}(|w_l|)$ for all the layers. The experimental results reported in \cref{fig_ablation_4} shows that the initialization does not play an important role for network ternarization in our case. The reason of that may comes to twofolds: 1) on one hand, all the layer-wise ternarization thresholds are initialized with small values where the difference is not significant. 2) on the other hand, all the thresholds are fully trainable which will mitigate the difference during training.

\subsection{Performance on ImageNet dataset}
Beyond the ablation studies we performed on the CIFAR-10 dataset, we also conduct the experiment on large scale ImageNet dataset with ResNet-18/34/50 (type-b residual connection) network structures. The experimental results are listed in \cref{table:resnet_imagenet} together the methods adopted in related works. Since for the realistic case that neural network operating on the specifically designed hardware, it is expected that all the layers are ternarized. The results shows that, our result can achieve the state-of-the-art results. The layer-wise thresholds are initialized as $\delta_l = 0.1\times |max(\bm{w}_l)|$. We use the full-precision pretrained model for weight initilization as described in \cref{fig_train_flow}. The learning rate starts from 1e-4, then change to 2e-5, 4e-6, 2e-6 at epoch 30, 40, 45 correspondingly.


\section{Conclusion and future works}
In this work, we have proposed a neural network ternarization method which incorporate thresholds as trainable parameter within the network inference path, thus both weights and thresholds are updated through back-propagation. Furthermore, we explicitly discuss the importance of straight-through-estimator design for approximating the gradient for staircase function. In general, our work is based on the assumption that the weight of deep neural network is tend to following Gaussian distribution. It turns out that such assumption somehow successfully returns a abstract model for network ternarization purpose.



{\small
\bibliographystyle{ieeetr}
\bibliography{./reference}
}

\end{document}